\documentclass{article} 
\usepackage{gmas_iclr2022_conference,times}
\usepackage{multirow}
\usepackage{enumitem}
\usepackage{xcolor}
\usepackage{marvosym}
\usepackage{xspace}
\usepackage{color}
\usepackage{colortbl}
\usepackage{graphicx}
\usepackage{booktabs}
\usepackage{wrapfig}
\usepackage[colorlinks=true,
            linkcolor={red!50!black},       
            urlcolor={green!50!black},
            citecolor={blue!50!black},
            ]{hyperref}
\usepackage[linesnumbered,ruled,vlined]{algorithm2e}

\newcommand{\gr}{\rowcolor[gray]{.95}}

\newcommand{\maxpressure}{\textsc{MaxPressure}\xspace}
\newcommand{\fixedtime}{\textsc{Fixedtime}\xspace}
\newcommand{\sotl}{\textsc{SOTL}\xspace}
\newcommand{\individualrl}{\textsc{Individual RL}\xspace}
\newcommand{\metalight}{\textsc{MetaLight}\xspace}
\newcommand{\presslight}{\textsc{PressLight}\xspace}
\newcommand{\colight}{\textsc{CoLight}\xspace}
\newcommand{\generalight}{\textsc{GeneraLight}\xspace}
\newcommand{\base}{\textsc{Base}\xspace}
\newcommand{\baseraw}{\textsc{Base+raw}\xspace}
\newcommand{\baseper}{\textsc{Base+shr}\xspace}
\newcommand{\basetem}{\textsc{Base+spe}\xspace}
\newcommand{\mtlight}{\textsc{MTLight}\xspace}

\newcommand{\hangzhou}{$\mathcal{D}_{Hangzhou}$\xspace}
\newcommand{\jinan}{$\mathcal{D}_{Jinan}$\xspace}
\newcommand{\newyork}{$\mathcal{D}_{NewYork}$\xspace}
\newcommand{\shenzhen}{$\mathcal{D}_{Shenzhen}$\xspace}
\newcommand{\perLatentState}{$\mathrm{\mathbf{o}_{t}^{shr}}$\xspace}
\newcommand{\temLatentState}{$\mathrm{\mathbf{o}_{t}^{spe}}$\xspace}
\newcommand{\realflow}{\textit{Real}\xspace}
\newcommand{\synflow}{\textit{Synthetic}\xspace}

\usepackage{amsmath,amsfonts,bm}

\def\eqref#1{equation~\ref{#1}}

\def\1{\bm{1}}

\DeclareMathAlphabet{\mathsfit}{\encodingdefault}{\sfdefault}{m}{sl}
\SetMathAlphabet{\mathsfit}{bold}{\encodingdefault}{\sfdefault}{bx}{n}

\usepackage{hyperref}
\usepackage{url}

\title{MTLight: Efficient Multi-Task Reinforcement Learning for Traffic Signal Control}

\author{Liwen Zhu \\
Peking University\\
\texttt{liwenzhu@pku.edu.cn} \\
\And
Peixi Peng \\
Peking University\\
\texttt{pxpeng@pku.edu.cn} \\  
\And
Zongqing Lu\\
Peking University\\
\texttt{zongqing.lu@pku.edu.cn} \\
\And
Yonghong Tian \\
Peking University\\
\texttt{yhtian@pku.edu.cn} \\
}

\iclrfinalcopy 
\begin{document}

\maketitle

\begin{abstract}
Traffic signal control has a great impact on alleviating traffic congestion in modern cities. Deep reinforcement learning (RL) has been widely used for this task in recent years, demonstrating promising performance but also facing many challenges such as limited performances and sample inefficiency. To handle these challenges, \mtlight is proposed to enhance the agent observation with a latent state, which is learned from numerous traffic indicators. Meanwhile, multiple auxiliary and supervisory tasks are constructed to learn the latent state, and two types of embedding latent features, the task-specific feature and task-shared feature, are used to make the latent state more abundant. Extensive experiments conducted on CityFlow demonstrate that \mtlight has leading convergence speed and asymptotic performance. We further simulate under peak-hour pattern in all scenarios with increasing control difficulty and the results indicate that \mtlight is highly adaptable.
\end{abstract}

\maketitle

\section{introduction}
\label{sec:introduction}

Traffic signal control aims to coordinate traffic signals across intersections to improve the traffic efficiency of a district or a city, which plays an important role in efficient transportation. 
Most conventional methods aim to control traffic signals by fixed-time \cite{koonce2008traffic} or hand-crafted heuristics \cite{kouvelas2014maximum}, which heavily rely on expert knowledge and in-depth excavation of regional historical traffic, making it difficult to migrate. 
Recently, deep reinforcement learning (DRL) based methods \cite{guo2021urban,jintao2020learning,pan2020spatio,he2020spatio,tong2021combinatorial,wang2020deep,gu2020exploiting,liu2021urban,xu2021hierarchically,zhang2021periodic} employ a deep neural network to control an intersection where the network is learned by directly interacting with the environment. However, due to the plenty of traffic indicators (number of vehicles, queue length, waiting time, speed, etc.), complex observation and the dynamic environment, the problem is challenging and remains unsolved.

Since the observation, reward and dynamics of each traffic signal are closely related to others, hence optimizing traffic signal control in a large-scale road network is naturally  modeled as a multi-agent reinforcement learning (MARL) problem. Most exiting works \cite{wei2019presslight,zhang2020generalight,chen2020toward,zheng2019learning} are proposed to learn the policy of each agent only conditioned on the raw observations of the intersection, while ignoring the help of the global state, which is accessible in smart city. 
As stated in \cite{zheng2019diagnosing}, different metrics have a considerable impact on the traffic signal control task. Hence, the observation design of agent should not only involve the raw observations of the intersection, but also the global state. A good agent observation design could make full use of samples, and improves not only the policy performance but also  the sample efficiency.
However, there are a huge amount of traffic indicators or metrics in the global state,
and it is hard to subjectively design suited and non-redundant agent observation among these indicators. On one hand, an overly concise observation design could not adequately and comprehensively represent the state characteristics and therefore affects the accuracy of the estimation of state transition and as well as influencing action selection. In contrast, if an overly complex combination of metrics is used as an observation, the 
weights of different metrics are difficult to precisely define, and it may cause data redundancy and dimension explosion, which will not only increase the computational consumption, but also make the agent hard to learn.

\begin{wrapfigure}{r}{0.34\textwidth}
    \centering
    \setlength{\abovecaptionskip}{5pt}
    \includegraphics[width=0.34\textwidth]{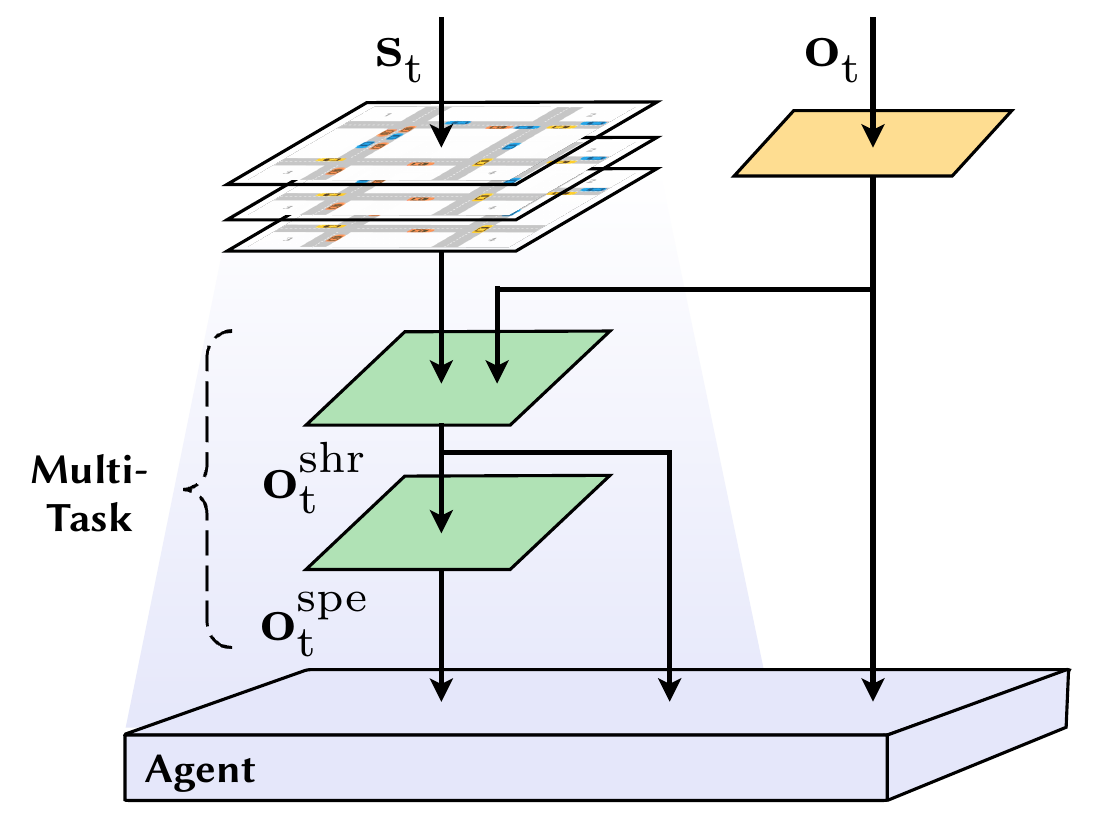}
    \caption{Multi-Task module forms task-shared and task-specific latent states to enhance the agent observation.} 
    \vspace{-0.35cm}
    \label{fig:motivation}
\end{wrapfigure}

In order to provide an adequate representation of the traffic signal control task, the latent state is introduced. Specifically, the raw observation is identical to the intersection, which consists of several variables with concrete semantic meanings (i.e., the number of vehicles on each incoming lane and current signal phase). Then, the raw observation is enhanced by the latent space. To learn the latent space from the global state, multiple auxiliary and supervisory tasks are constructed, which are related to traffic signal control. That is, several statistics of global state history are taken as inputs, a RNN-based network is employed firstly, and several branches are introduced subsequently to predict multiple types of statistics of the global state, such as the flow distribution and the travel time distribution, respectively. To make the latent space more abundant, two types of embedding features are extracted: the task-specific feature and task-shared feature. The former is extracted by the task-specific branch and represents the task-driven information, while the later is from the task-shared layer and could express more general underlying 
characteristics. Hence, they are complementary to each other and are both used to enhance the raw observation. Finally, conditioned on the enhanced observation, the policy is learned by DRL \cite{mnih2015human}. Note that the  multiple tasks are learned simultaneously with the DRL, which makes the latent space more adaptive to the policy learning.

\section{Problem Statement}
\label{sec:problem_statement}

\subsection{Problem Definition}
\label{sec:problem_definition}


We consider a multi-agent traffic signal control problem, the task is modeled as a Markov Game \cite{littman1994markov}, which can be denoted by a tuple $\mathcal{G}=<\mathcal{N},\mathcal{S}, \mathcal{A}, \mathcal{O}, \mathcal{P}, \mathcal{R}, \mathcal{H}, \gamma>$. $\mathcal{N} \equiv\{1, \ldots, n\}$ is a finite set of agents, and each intersection in the scenario is controlled by an agent. $\mathcal{S}$ is a finite set of global state space. $\mathcal{A}$ denotes the action space for an individual agent. The joint action $\boldsymbol{a} \in \mathbf{A} \equiv \mathcal{A}^{n}$ is a collection of individual actions $\left[a_{i}\right]_{i=1}^{n}$. At each timestep, each agent $i$ receives an observation $o_{i} \in \mathcal{O}$, selects an action $a_{i}$, results in the next state $s^{\prime}$ according to the transition function $\mathcal{P}\left(s^{\prime} \mid s, \boldsymbol{a}\right)$ and a reward $r=\mathcal{R}(s, \mathbf{a})$ for each agent. $\mathcal{H}$ is the time horizon and $\gamma \in[0,1)$ is the discount factor.

\subsection{Agent Design}
\label{sec:agent_design}

Each intersection in the system is controlled by an agent. In the following, we introduce the state design, action design and reward design of the RL agent.

\begin{itemize}[leftmargin=15pt]
\item
\textbf{Observation.} Our primitive observation consists of two parts: (1) the number of vehicles on each incoming lane $\mathbf{f}_t^v$; (2) current signal phase $\mathbf{f}_t^s$. Both of them can be obtained directly from the simulator, the concepts are described in detail in Section \ref{sec:preliminaries}. The raw observation of agent $i$ is defined by
\begin{align}
    o_{i} = \{ \mathbf{f}_t^v, \mathbf{f}_t^s \},
\end{align}
where $\mathbf{f}_t^v = \{{V}_{l_{1}^{in}}, {V}_{l_{2}^{in}}, \ldots, {V}_{l_{m}^{in}} \}$ and ${l}^{in} = \{l_{1}^{in}, \ldots, l_{m}^{in}\}$ is a finite set of incoming lanes in the intersection. Current signal phase $\mathbf{f}_t^s = {p}_{k}, k \in {1, \ldots, K}$, and $K$ is the total number of phases. Each phase $p$ is represented as a one-hot vector. Our goal is to learn latent space to enhance the raw observation to make better use of the sample.

\item
\textbf{Action.} The action of each agent is to choose the phase for the next time interval. Note that the phases may organize in a sequential way in reality, while directly selecting a phase makes the traffic control plan more flexible. Action of agent $i$ is defined by
\begin{align}
    a_{i} = \{ \mathbf{f}_t^s\},
\end{align}
where $\mathbf{f}_t^s = {p}_{k}, k \in {1, \ldots, K}$.

\item
\textbf{Reward.} We define the reward as the negative of the queue length on incoming lanes, which is generally accepted and reasonable in previous work \cite{zheng2019diagnosing,huang2021modellight,zang2020metalight,zheng2019learning,wei2019colight}. 
Reward of agent $i$ is defined by
\begin{align}
    r_{i} =  -\sum^{M}_{m} q_{l^{in}_{m}},
\end{align}
where $q_{l^{in}_{m}}$ is the queue length on incoming lane $l^{in}_{m}$.
\end{itemize}

\section{method}
\label{sec:method}
\begin{figure*}[t]
 		\centering
 		\includegraphics[width=0.95\textwidth]{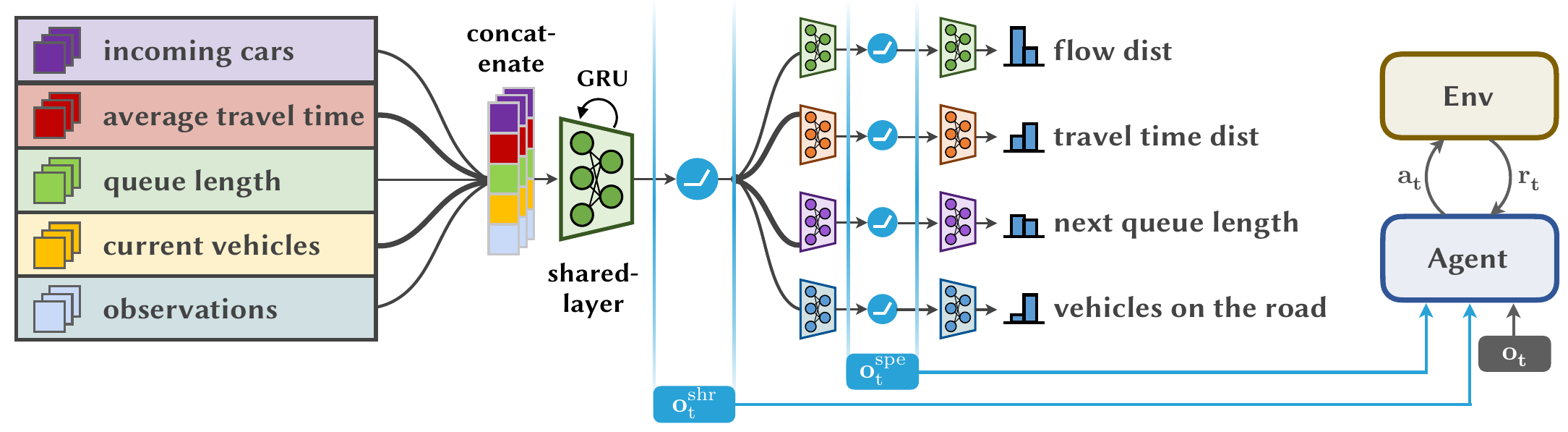} 
 		\vspace{-0.2cm}
 		\caption{MTLight consists of a multi-task network and a policy network. RL agent is augmented with a task-shared latent state \perLatentState and a task-specific latent state \temLatentState.}
 		\label{fig:framework}
 		\vspace{-0.4cm}
\end{figure*}

In this section, we will introduce the main modules of our proposed method \mtlight, which focuses on learning task-related task-shared latent state and task-specific latent state by introducing an auxiliary Multi-Task network to help policy learning. The whole process of \mtlight is described in Algorithm \ref{alg:train}, and the framework of \mtlight is shown in Fig. \ref{fig:framework}.

\mtlight consists of a Multi-Task network and an agent network. For the latter, Deep Q-Network (DQN) \cite{mnih2015human} is employed  as function approximator to estimate the Q-value function, which is consistent with the previous methods \cite{chen2020toward,wei2019colight,wei2019presslight,zheng2019learning,wei2018intellilight}. The Multi-Task module adopts a hard parameter sharing paradigm \cite{caruana1997multitask}, which generally applied by sharing the hidden layers between all tasks, while keeping several task-specific output layers.

\subsection{Multi-Task Learning for Latent State}

For each agent, its raw observation includes the number of vehicles $\mathbf{f}_t^v$ and the current signal phase $\mathbf{f}_t^s$. 
Besides, several information from the global state is given, such as: the number of incoming cars in the last $\tau$ steps, denoted as $\mathbf{f}_{t-\tau:t}^c = [\mathbf{f}_{t-\tau}^c, \mathbf{f}_{t-\tau+1}^c, \ldots, \mathbf{f}_{t}^c]$, the average travel time during the past $\tau$ steps, denoted as $\mathbf{f}_{t-\tau:t}^{tr} = [\mathbf{f}_{t-\tau}^{tr}, \mathbf{f}_{t-\tau+1}^{tr}, \ldots, \mathbf{f}_{t}^{tr}]$, the queue length during the past $\tau$ steps, denoted as $\mathbf{f}_{t-\tau:t}^{q} = [\mathbf{f}_{t-\tau}^{q}, \mathbf{f}_{t-\tau+1}^{q}, \ldots, \mathbf{f}_{t}^{q}]$, the current vehicles during the past $\tau$ steps, which is denoted as $\mathbf{f}_{t-\tau:t}^{vr} = [\mathbf{f}_{t-\tau}^{vr}, \mathbf{f}_{t-\tau+1}^{vr}, \ldots, \mathbf{f}_{t}^{vr}]$.

\vspace{0.2cm}
The Multi-Task module includes the following four tasks:
\begin{enumerate}[leftmargin=20pt]

    \item 
    \textbf{Flow distribution approximation.}
    We use $\mathcal{T}_{flow}$ to denote the traffic distribution estimation task, i.e., to predict the mean $\mu_{f}$ and variance $\sigma_{f}^{2}$ of flow arrival rate from start up to the time step $t$. The task could be denoted as: 
    \begin{align}
        (\mu_{f}, \sigma_{f}^{2}) \leftarrow  [\mathbf{f}_t^v, \mathbf{f}_t^s, \mathbf{f}_{t-\tau:t}^c, \mathbf{f}_{t-\tau:t}^{tr}, \mathbf{f}_{t-\tau:t}^{q}, \mathbf{f}_{t-\tau:t}^{vr}]. 
    \end{align}
    
    \item
    \textbf{Travel time distribution approximation.}
    We use $\mathcal{T}_{travel}$ to denote the travel distribution estimation task, i.e., to predict the mean $\mu_{tr}$ and variance $\sigma_{tr}^{2}$ of average travel time of vehicles that have completed the trip from start up to the time step $t$:
    \begin{align}
       (\mu_{tr}, \sigma_{tr}^{2}) \leftarrow [\mathbf{f}_t^v, \mathbf{f}_t^s, \mathbf{f}_{t-\tau:t}^c, \mathbf{f}_{t-\tau:t}^{tr}, \mathbf{f}_{t-\tau:t}^{q}, \mathbf{f}_{t-\tau:t}^{vr}]. 
    \end{align}
    
    \item
    \textbf{Next queue length approximation.}
    We use $\mathcal{T}_{queue}$ to denote the next queue length estimation task, i.e., to predict the average number $q$ of vehicles in queue at the next step:
    \begin{align}
        q \leftarrow [\mathbf{f}_t^v, \mathbf{f}_t^s, \mathbf{f}_{t-\tau:t}^c, \mathbf{f}_{t-\tau:t}^{tr}, \mathbf{f}_{t-\tau:t}^{q}, \mathbf{f}_{t-\tau:t}^{vr}]. 
    \end{align}
    
    \item
    \textbf{Vehicles on the road approximation.}
    We use $\mathcal{T}_{vehicles}$ to denote the vehicles on the road approximation task, i.e., to predict the number of vehicles $ V^{r}$ existing in the system:
    \begin{align}
        V^{r} \leftarrow [\mathbf{f}_t^v, \mathbf{f}_t^s, \mathbf{f}_{t-\tau:t}^c, \mathbf{f}_{t-\tau:t}^{tr}, \mathbf{f}_{t-\tau:t}^{q}, \mathbf{f}_{t-\tau:t}^{vr}]. 
    \end{align}
Note that vehicles that have completed the trips or have not yet entered the road network do not belong to these. 
\end{enumerate}

The above tasks act auxiliary tasks to learn the latent space. Since the numbers of $\mathbf{f}_{t-\tau:t}^c$,  $\mathbf{f}_{t-\tau:t}^{tr}$, $\mathbf{f}_{t-\tau:t}^{q}$, $\mathbf{f}_{t-\tau:t}^{vr}$ have different scales and their dimensions are different with $\mathbf{f}_t^v$ and $\mathbf{f}_t^s$, four independent linear layers and ReLU functions are employed firstly to scale them respectively: 
\begin{align}
    \mathbf{h}^{c} = {ReLU}(\mathbf{W}_{1} \mathbf{f}_{t-\tau:t}^{c}+\mathbf{b}_{1}),
    \ 
    \mathbf{h}^{tr} = {ReLU}(\mathbf{W}_{2} \mathbf{f}_{t-\tau:t}^{tr}+\mathbf{b}_{2}), \\
    \mathbf{h}^{q} = {ReLU}(\mathbf{W}_{3} \mathbf{f}_{t-\tau:t}^{q}+\mathbf{b}_{3}),
    \ 
    \mathbf{h}^{vr} = {ReLU}(\mathbf{W}_{4} \mathbf{f}_{t-\tau:t}^{vr}+\mathbf{b}_{4}).
\end{align}


Then a linear layer and ReLU function is used to calculate the hidden state after concatenating all embedded inputs: 
\begin{align}
    \mathbf{H}_{t} = {ReLU}(\mathbf{W}_{} (\mathbf{f}_t^v, \mathbf{f}_t^s, \mathbf{h}^{c}, \mathbf{h}^{tr}, \mathbf{h}^{q}, \mathbf{h}^{vr})+\mathbf{b}_{}).
\end{align}
Based on $\mathbf{H}_{t}$, a task-shared network module is used to generate its task-shared latent feature (\perLatentState, also called \textit{apparent state}). Then, 4 independent branches are introduced for each task and calculate task-specific latent feature (\temLatentState, also called \textit{mental state}) from \perLatentState. The specific implementation of network architecture is listed in the supplementary.

We use a single latent variable model to extract hierarchical latent features, which follows insights by \cite{zhao2017learning}. That is, the \textit{mental state} is output of the shared-layer after GRU in Multi-Task network and could express more general underlying 
characteristics. In contrast, the \textit{apparent state} is the the concatenation of the output of the task-specific layer and  represents the task-driven information. In other words, the \textit{mental state} is more coarse-grained, while \textit{apparent state} is more fine-grained. Hence, they are complementary to each other and both used in our method.

\subsection{Policy with Latent State}

With the help of latent state, the agent observation is enhanced from $\mathrm{\mathbf{o}_t}$ to $(\mathrm{\mathbf{o}_t},\mathrm{\mathbf{o}_{t}^{shr}},\mathrm{\mathbf{o}_{t}^{spe}})$. For the policy $\pi^{\theta}$, the objective is to maximize the cumulative reward:
\begin{align}
    \max\limits_{\theta}J(\theta)=\mathbb{E}_{\substack{a_t \sim \pi^\theta(a_t \mid \mathrm{\mathbf{o}_t},\mathrm{\mathbf{o}_{t}^{shr}}. \mathrm{\mathbf{o}_{t}^{spe}})}}\sum\limits_{t=0}^{\mathcal{H}-1}\gamma^{t}r_{t+1}.
    \label{eq:RL}
\end{align}

An agent that maximises Eq. \ref{eq:RL} acts optimally under uncertainty and is called \textit{Bayes-optimal} \cite{ghavamzadeh2015bayesian}, assuming we treat the knowledge over related tasks as our epistemic prior about the environment. Multi-Task module minimizes the complexity of the model and give informative priors to the model. Besides, it can minimize the representation bias in a way that push the learning algorithm to find a solution on a smaller area of representations on the intersection rather than on a large area of a single task. This incentivises a faster and better convergence.


\section{experiment}
\label{sec:experiment}

We conduct the experiments on CityFlow \cite{zhang2019cityflow}, an city-level open-source simulation platform for traffic signal control. The simulator is used as the environment to provide state for traffic signal control, the agents execute actions by changing the phase of traffic lights, and the simulator returns feedback.

Please refer to Appendix \ref{sec:road_networks} and Appendix \ref{sec:flow_configurations} for the detailed settings of road network and traffic flow configuration. Baselines are described in detail in Appendix \ref{sec:baselines}.

\subsection{Performance Comparison}

\begin{figure*}[t]
 		\centering
 		\includegraphics[width=1.0\textwidth]{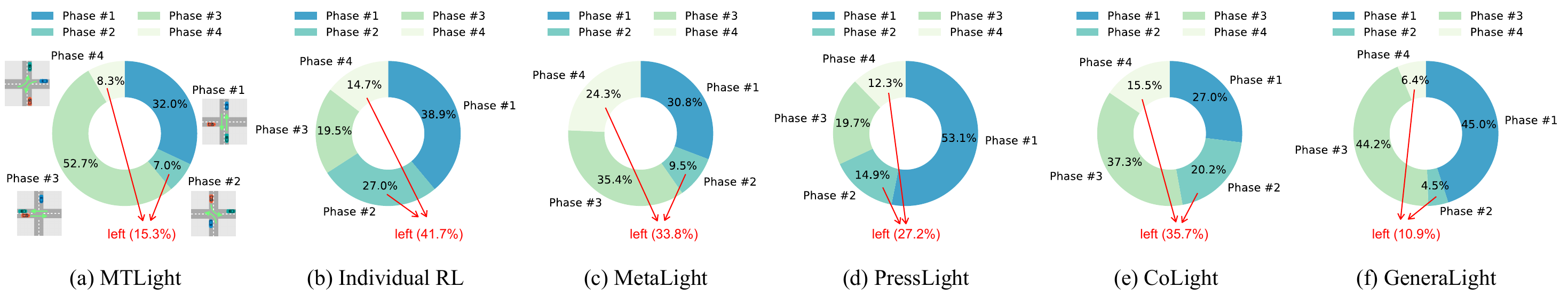} 
 		\vspace{-0.6cm}
 		\caption{Illustration of strategies for all RL methods under \realflow configuration in Hangzhou.}
 		\label{fig:action_distribution}
 		\vspace{-0.4cm}
\end{figure*} 

\renewcommand\tabcolsep{2.8pt}  

    \begin{table*}[h]
    \centering
    \small
    \setlength{\belowcaptionskip}{6pt}
    \caption{Overall performance comparison on Hangzhou, Jinan, New York and Shenzhen under \realflow and \synflow configurations. Average travel time is reported in the unit of second. "Mean" in the last column shows the average performance of the scenarios shown in the previous 8 columns.}
    \label{tab:performance_1}
    \begin{tabular}{l | rr |  rr | rr | rr | r}
    \toprule
    \multirow{3}{*}{\textbf{Model}}
    & \multicolumn{2}{c|}{\textbf{Hangzhou}}
    & \multicolumn{2}{c|}{\textbf{Jinan}} 
    & \multicolumn{2}{c|}{\textbf{Newyork}} 
    & \multicolumn{2}{c|}{\textbf{Shenzhen}} 
    & \multirow{3}{*}{\textbf{Mean}} \\
    
    \cmidrule(lr){2-3} \cmidrule(lr){4-5} \cmidrule(lr){6-7} \cmidrule(lr){8-9}
    & \textbf{real} & \textbf{syn\_peak} & \textbf{real} & \textbf{syn\_peak} & \textbf{real} & \textbf{syn\_peak}  & \textbf{real} & \textbf{syn\_peak} & \\
    \midrule
    \maxpressure & 416.82 & 2320.65 & 355.12 & 1218.13 & 380.42 & 1481.48 & \textbf{389.45} & 1387.87 & 1387.87 \\
    \fixedtime & 718.29 & 1787.58 & 814.09 & 1739.69 & 1849.78 & 2086.59 & 786.54 & 1845.03 & 1453.45 \\
    \sotl & 1209.26 & 2062.49 & 1453.97 & 1991.03 & 1890.55 & 2140.15 & 1376.52 & 2098.09 & 1777.76 \\
    \midrule
    \individualrl & 743.00 & 1819.57 & 843.63 & 1745.07 & 1867.86 & 2100.68 & 769.47 & 1845.34 & 1466.83 \\
    \metalight & 480.77 & 1576.32 & 784.98 & 1854.38 & 261.34 & 2145.49 & 694.83 & 2083.26 & 1235.17 \\
    \presslight & 529.64 & 1754.09 & 809.87 & 1930.98 & 302.87 & 1846.76 & 639.04 & 1832.76 & 1205.75 \\
    \colight & 297.89 & 1077.29 & 511.43 & 1217.17 & \textbf{159.81} & 1457.56 & 438.45 & 1367.38 & 815.87 \\
    \generalight & 335.18 & 1574.93 & 585.89 & 1616.28 & 1208.73 & 1686.49 & 792.22 & 1574.10 & 1171.73 \\
    \midrule 
    \base & 705.85 & 1718.37 & 808.28 & 1703.21 & 903.82 & 2097.84 & 728.49 & 1937.45 & 1325.41 \\
    \baseraw & 684.34 & 1845.92 & 623.94 & 1835.45 & 592.34 & 1934.04 & 703.56 & 1845.32 & 1258.11 \\
    \baseper & 313.28 & 1146.79 & 499.88 & 1325.27 & 463.15 & 1416.65 & 438.69 & 1371.53 & 871.91 \\
    \basetem & 431.55 & 1446.63 & 517.09 & 1430.96 & 431.65 & 1669.61 & 684.83 & 1442.35 & 1006.83 \\
    \mtlight & \textbf{161.24} & \textbf{1011.67} & \textbf{346.93} & \textbf{1176.02} & 209.46 & \textbf{1394.15} & 402.57 & \textbf{1284.93} & \textbf{748.37} \\
    \bottomrule
    \end{tabular}
\end{table*}

\begin{wrapfigure}{r}{0.4\textwidth}
    \centering
    \setlength{\abovecaptionskip}{5pt}
    \includegraphics[width=0.42\textwidth]{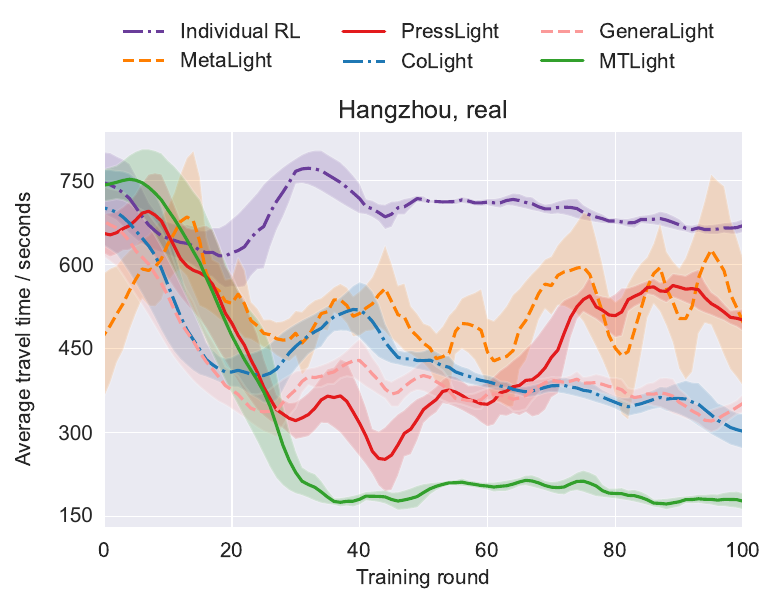}
    \vspace{-0.5cm}
    \caption{Performance of RL methods under real configurations.} 
    \vspace{-0.35cm}
    \label{fig:hangzhou_real}
\end{wrapfigure}

Tab. \ref{tab:performance_1} lists the comparative results, and it is evident that: 1) In general, RL methods perform better than conventional methods, and it indicates the advantage of the RL. Moreover, \mtlight is outperforms other methods in almost all cities and flow configurations, which demonstrates the effectiveness of the method. 2) \mtlight shows good generalization for different scenarios and configurations. For example, \maxpressure performs well in \hangzhou with the \realflow, while under the \synflow traffic conditions, \maxpressure shows significantly worse than other methods. In contrast, \mtlight can not only achieve good performance under diverse configurations of \hangzhou, but also shows great stability. 3) \mtlight outperforms \individualrl, \metalight and \presslight with 693.46, 461.80 and 432.38, respectively. The reason is that they learn the traffic light's policy only using its observation and ignore the influence of the neighbors, while \mtlight considers the neighbors as the latent part of the environment to help learning. 4) The neighbor's information is modeled in \colight and \generalight can adapt to a variety of flows, they both perform well. While results of \mtlight is superiors to them in multiple scenarios, resulting mean 42.5 and 398 improvement. Compared to them, \mtlight benefits from prior knowledge learned from Multi-Task network to make more accurate decisions.

\begin{wrapfigure}{r}{0.4\textwidth}
    \centering
    \setlength{\abovecaptionskip}{5pt}
    \includegraphics[width=0.42\textwidth]{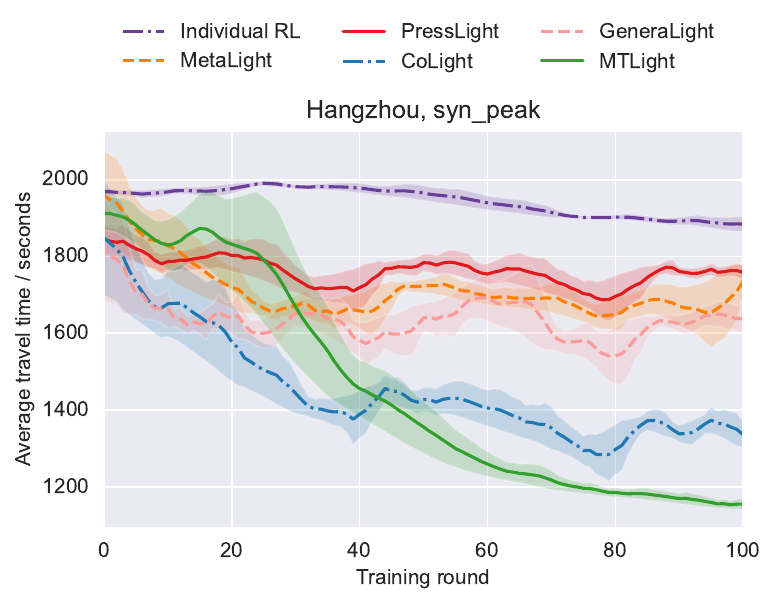}
    \caption{Performance of RL methods under synthetic peak configurations.} 
    \vspace{-0.35cm}
    \label{fig:hangzhou_syn_peak}
\end{wrapfigure}

Fig. \ref{fig:hangzhou_real} shows the performances of all RL methods of \hangzhou under \realflow traffic pattern, and it is obvious that \mtlight converges faster and has better asymptotic performance. Fig. \ref{fig:hangzhou_syn_peak} shows the performances of all RL methods of \hangzhou under \synflow traffic pattern, we can conclude that \mtlight converges quickly and learns effectively during the peak hour, while the other method have only a weak boost during training.

Fig. \ref{fig:turning} and Tab. \ref{tab:turning} illustrates the turning statistics of vehicle routes. Take \hangzhou \realflow as an example, the frequency of turning left and going straight is 14\% and 86\% respectively (turning right are not considered because they are free from the control by lights). Fig. \ref{fig:action_distribution} shows the percentage of each phase of RL methods, we can find: 1) The total left-turn phase of \mtlight accounts for 15.3\%, which is highly consistent with the left-turn frequency of 14\%, which indicates that the strategy is interpretable. 2) The \generalight left-turn ratio of 10.9\% is also close, but because it has an excessive proportion of straight phases, it may cause left-turn vehicles to be stranded, resulting in increased travel time. 3) \individualrl tends to consider phase 1 and 2, which account for as much as 65.9\%, \metalight prefers to go straight, \presslight is eccentric to phase 1, and \colight assigns a relatively even distribution to each phase, rather than aligning with the traffic flow direction. These all demonstrate the limitations of other RL methods in multi-agent environments, while \mtlight can learn more stable strategies by introducing task-shared and task-specific latent states.

\subsection{Ablations}
To better validate the contribution of each component, three variants of \mtlight are evaluated under a variety of scenarios, as shown in Tab. \ref{tab:performance_1}.
\begin{itemize}[leftmargin=15pt]
\item
\textbf{\base} only keeps the policy network and removes the Multi-Task network.

\item
\textbf{\baseraw} only keeps the policy network and discards Multi-Task network, but directly uses the original input of Multi-Task module as part of the observation. 

\item
\textbf{\baseper} retains the Multi-Task network and the policy, but only has task-shared latent state and removes task-specific latent state.

\item
\textbf{\basetem} retains the Multi-Task network and the policy. In contrast to \baseper, \basetem has only the task-specific latent state and removes the task-shared latent state.

\end{itemize}

Note that \mtlight contains the whole modules: policy network, Multi-Task network with both task-specific latent state and task-shared latent state.

The quantitative evaluation results are presented in Tab. \ref{tab:performance_1}. We can obtain the following findings:
1) Among these 4 models, the performance of \base is the worst. The reason is that it is hard to learn the effective policy independently in the multi-agent traffic signal control task, where the surrounding environment is changing dynamically, but \base has no sense of it. 2) Compared with the \base and \baseraw, the improvement of \baseper and \basetem demonstrate the effectiveness of the task-shared latent state \perLatentState and task-specific latent state \temLatentState respectively. \perLatentState reflects prior information that is constant over time with multiple related tasks , \temLatentState reflects prior information that is align with the latest changing trends, both of them help policy to make Bayesian optimal decisions. 3) The \perLatentState and \temLatentState are both effective because each of them is an efficient representations of environmental features. Compared to them, the superiority of \mtlight indicates \perLatentState and \temLatentState are complementary to each other. 
Overall, all of the proposed components contribute positively to the final results.

\section{conclusion}
\label{sec:conclusion}
We introduced \mtlight, an efficient Multi-Task reinforcement learning method for traffic signal control that can be scaled to complex multi-agent urban road networks of different scale. We showed that \mtlight's latent structure learns a hierarchical latent representations of related tasks, separating the task-shared and task-specific latent states. On several cities’ datasets we demonstrated that this latent representation inspired from related multiple tasks, and conditioning the policy on it, allows an agent to adapt to the complex environment. We conclude that maintaining prior approximations over related tasks helps compared to model-free approaches, especially when there is too much information in the environment and it cannot be fully expressed by artificial state design.

For the future, the latent prior could be learned from expert data prepared in advance using imitation learning techniques \cite{song2018multi}, or by using existing multi-agent algorithms to pre-train Multi-Task network.

\footnotesize
\bibliography{gmas_iclr2022_conference}
\bibliographystyle{gmas_iclr2022_conference}
\normalsize

\clearpage
\appendix
\section{Appendix}
You may include other additional sections here.


\renewcommand\tabcolsep{12pt}
\begin{table}[h]
    \normalsize 
    \centering
    \setlength{\belowcaptionskip}{6pt}
    \caption{Implementation details of \mtlight}
    \label{tab:implementation_details}
    \begin{tabular}{ll}
    \toprule 
    Items                           & Details                           \\
    \midrule
    Number of policy steps          & 3600                             \\ 
	Discount factor $\gamma$        & 0.95                              \\ 
	Policy $\epsilon$               & 0.1 $\rightarrow$ 0.01                                \\
	$\epsilon$ decay rate           & 0.995                                \\
	Policy Learning rate            & 0.005                              \\ 
	Policy minibatch                & 32                             \\ 
	task-shared latent space dim      & 5                             \\
	task-specific latent space dim       & 5                     \\
	task-shared latent state coef & 10                 \\
	task-specific latent state coef & 10                      \\
	\midrule
	Policy network                  & 2 hidden layers,                  \\
	architecture                    & 20 nodes each,                    \\
	                                & ReLU activations                  \\
	\midrule
	Policy network                  & RMSprop with learning rate 0.001   \\
	optimizer                       & and MSE loss                  \\
	\midrule
	                                & 5 MLP embedding layers ,           \\
	                                & 2 shared FC layers before GRU,           \\
	                                & GRU with hidden size 64,          \\ 
	Multi-Task architecture         & 1 shared FC layer after GRU,    \\
	                                & 4 task-specific FC layers,        \\
	                                & 4 output task layers                 \\
	                                & ReLU activations                  \\
	\midrule
	Multi-Task optimizer            & Adam with learning rate 0.01    \\
	                                & and MSE loss                  \\
    \bottomrule
    \end{tabular}
\end{table}

\section{related work}
\label{sec:related_work}

\subsection{Conventional and Adaptive Traffic Signal Control}
Most conventional traffic signal control methods are designed based on fixed-time signal control \cite{webster1958traffic}, actuated control \cite{chiu1992adaptive} or self-organizing traffic signal control \cite{chiu1993self,cools2013self,lowrie1990scats,svanes1981scat,hunt1981scoot}. These approaches rely on expert knowledge and often perform unsatisfactorily in complicated real-world situations. To solve this problem, several optimization-based methods \cite{roess2004traffic,varaiya2013max,kouvelas2014maximum} have been proposed to optimize average travel time, throughput, \textit{etc.}, which decide the traffic signal plans according to the observed data instead of the human prior. However, these approaches typically rely on strict assumptions which might not hold in the real-world cases \cite{webster1966traffic}. Furthermore, the optimization problems are usually hard to tract and require significant computing power in complex scenarios.

\subsection{RL-based Traffic Signal Control}
RL-based traffic signal control methods aim to learn the policy from interactions with the environment. Earlier studies use tabular Q-learning \cite{el2013multiagent,abdoos2013holonic,dusparic2009distributed,abdoos2011traffic} where the states in an environment are required to be discretized and low-dimensional. To address the unmanageable large or continuous state space, recent advances employ deep RL with more complex continuous state representations (like images or feature vectors) to map the high-dimensional states into actions.

Efforts have been made to design strategies that formulate the task as a single agent \cite{wei2018intellilight,mannion2016experimental,huang2021modellight,zang2020metalight,oroojlooy2020attendlight,jiang2021dynamic,rizzo2019time} or some isolated intersections \cite{zheng2019diagnosing,zheng2019learning,xiong2019learning,wei2019presslight,chen2020toward,oroojlooy2020attendlight,zhang2020generalight,zhang2020planlight}, i.e., each agent makes decision for its own. The above methods  are usually easy to scale, but they may have difficulty achieving globally optimal performance due to a lack of collaboration. To solve the problem, another way is to consider jointly modeling the action between learning agents with centralized optimization \cite{van2016coordinated,kuyer2008multiagent}. However, 
as the number of agents increases, joint optimization usually leads to  dimensional explosion, which has inhibited the widespread adoption of such methods to a large-scale traffic signal control. To overcome the difficulty, another type of methods are implemented in a decentralized manner, taking into account the collaboration between neighbors with appropriate reward and state design \cite{arel2010reinforcement,nishi2018traffic,wei2019colight,xu2021hierarchically}. Methods such as \cite{el2013multiagent,chu2019multi} add neighboring information into states, \cite{nishi2018traffic,wei2019colight,yu2020macar,guo2021urban} add neighbors' hidden features into states, and \cite{xu2021hierarchically} optimizes neighborhood travel time as an additional reward. However, simple concatenation of neighboring information is not reasonable enough because the influence of neighboring intersections is not balanced. Unlike the above methods that add neighbor information to the state, our method learns task-shared and task-specific latent states by constructing Multi-Task network.

\subsection{Multi-Task Learning}
Multi-Task Learning(MTL) \cite{caruana1997multitask} is a learning paradigm aims to jointly learn multiple related tasks so that the knowledge contained in a task can be leveraged by other tasks. Past works \cite{oh2017zero,zhang2021survey,ruder2017overview,ndirango2019generalization} have found that, by sharing a representation among related tasks and jointly learning all the tasks, better generalization can be achieved over independently learning each task. Constructing auxiliary tasks to help the main task is a branch of Multi-Task Learning. Reinforcement learning is known to be sample inefficient, transferring knowledge from other auxiliary tasks is a powerful tool for improving the learning efficiency \cite{jaderberg2016reinforcement,lin2019adaptive,lyle2021effect,tongloy2017asynchronous,bellemare2019geometric}. \cite{lin2019adaptive} combines different auxiliary tasks which provide gradient directions to speed up the training of the main reinforcement learning task. In comparison, our work aims to transfer knowledge from the task-related auxiliary tasks as a prior to the main reinforcement learning task, to ultimately boost the performance. Specifically, we model the Multi-Task network as a latent structure where the task-shared latent state is generated from early layers and the task-specific latent state is generated from deeper layers. This incentivies the policy to learn the Bayers-optimal behaviours: the policy can take into account its uncertainty over the comprehensive information when choosing actions. 

\subsection{Preliminaries}
\label{sec:preliminaries}

In this section, we first introduce some basic concepts related to traffic signal control (TSC) that have been widely recognized in previous work \cite{wei2019colight,zheng2019learning,zhang2020generalight,wei2019presslight,chen2020toward,zang2020metalight}. Note that the concepts can be easily generalized to other intersections with different structures.


\begin{figure*}[t]
 		\centering
 		\includegraphics[width=0.6\textwidth]{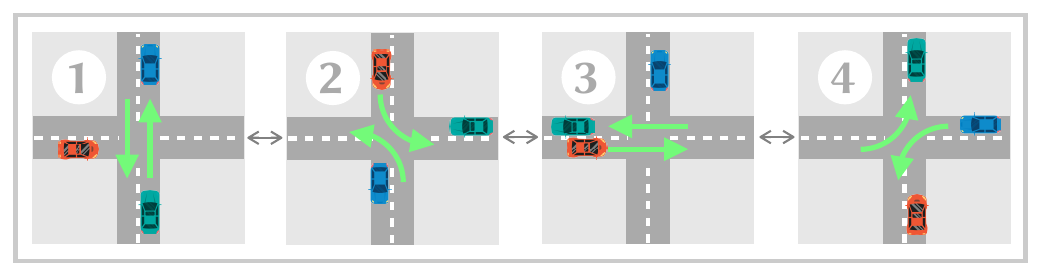} 
 		\vspace{-0.3cm}
 		\caption{Illustration of phase.}
 		\label{fig:phase_illustration}
 		\vspace{-0.35cm}
\end{figure*}

\begin{itemize}[leftmargin=15pt]

\item
\textbf{Incoming/Outgoing Lanes.} The incoming lanes refer to the lanes where the vehicles are about to enter the intersection. It usually contains three basic types: "left-turn", "straight" and "right-turn" from inner to outer. The outgoing lanes refer to the lanes where the vehicles are about to leave the intersection.

\item
\textbf{Roadnet.} A roadnet is a part of a dataset that represents an area of a city. A roadnet consists of signalized intersections, unsignalized intersections, and lanes connecting the intersections. Generally, the lane lengths, number of lanes and relative locations of intersections vary from one roadnet to another.

\item
\textbf{Phase.} Phase is a controller timing unit associated with the control of one or more movements, representing the permulation and combination of different traffic flows. The 4-phase setting is the most common configuration in reality, illustrated in Fig. \ref{fig:phase_illustration}, but the number of phases can vary due to different intersection topologies (3-way, 5-way intersections, etc.).

\item
\textbf{Queue Length.} Queue length is the number of vehicles waiting at an intersection due to a red light. Vehicles on the incoming lane with a speed of less than 0.1m/s are considered to be waiting.

\item
\textbf{Average Travel Time.} The travel time of a vehicle is the time discrepancy between entering and leaving a particular area. Average travel time of all vehicles in a road network is the most frequently used measure to evaluate the performance of traffic signal control \cite{wei2019colight,wei2019presslight,zhang2020generalight,chen2020toward,zheng2019learning}.

\item
\textbf{Flow Distribution.} Flow distribution is the distribution of traffic entering the road network, which is generally expressed by the arrival rate of vehicles, i.e., the volume of traffic entering the road network per unit time.

\item
\textbf{Vehicles on Road.} Vehicles on road indicate the running vehicle, i.e., vehicles that have entered the road network and have not reached the end point. Vehicles on road can represent the real-time load on the road network.

\end{itemize}

\section{Algorithm}

The algorithm is shown in Alg. \ref{alg:train}.
\begin{algorithm}[t]
\DontPrintSemicolon
\caption{Training Process of \mtlight}
\label{alg:train}
\SetKwInOut{Input}{\textbf{Input}}\SetKwInOut{Output}{\textbf{Output}}
\KwIn{
Roadnet file; traffic flow file; number of training episodes $E$; frequency of updating policy $t_p$; frequency of updating multi-task network $t_m$; total simulate time $T$
}
\KwOut{
Set of optimized parameters for the intersections; optimized parameter for the multi-task network
}
\BlankLine
Initialize task-shared and task-specific latent state \perLatentState, \temLatentState \\
Initialize policy replay buffer $\mathcal{B}^{\pi}$ \\
Initialize policy $\pi^{\theta}$ and multi-task network $\mathbf{M}^{\phi}$\\
Initialize reward of each agent $\{r_{i} \mid i \in 1, \ldots, n \}$ \\
\For{episode $\longleftarrow$ 1, 2, \dots, $E$}
{
    \For{ step t $\longleftarrow$ 1, 2, \dots, $T$}
    {
        Collect original observations for all agents \\
        Add task-shared \perLatentState and task-specific \temLatentState latent state to the observations \\
        \For{ agent i $\longleftarrow$ 1, 2, \dots, n}
        {
            Select action according to $\pi^{\theta}$ \\
        }
        Employ joint action $\boldsymbol{a}$ to the environment \\
        Get new observations and environmental reward \\
        Collect trajectories to replay buffer $\mathcal{B}^{\pi}$ \\
        Get multi-task network input $\mathbf{f}_t^v, \mathbf{f}_t^s, \mathbf{f}_{t}^{c}, \mathbf{f}_{t}^{tr}, \mathbf{f}_{t}^{q}, \mathbf{f}_{t}^{vr}$ from the environment \\
        Predict results using multi-task network $\mathbf{M}^{\phi}$ \label{step_a} \\ 
        Get task-shared \perLatentState and task-specific \temLatentState latent state from $\mathbf{M}^{\phi}$ \\
        Calculate statistics from 0 up to $t$ as supervised signal \label{step_b} \\
        \If{t = $t_p$}
        {
            Train policy $\pi^{\theta}$ by maximizing reward in Eq. \ref{eq:RL} \\
            Clean up $\mathcal{B}^{\pi}$
        }
        \If{t = $t_m$}
        {
            Calculate loss from the results of step \ref{step_a} and step \ref{step_b} \\
            Train multi-task network $\mathbf{M}^{\phi}$ \\
        }
        \If{t = $T$}
        {
            Collect the average total travel time of all vehicles as criteria \\
        }
    }
}
\end{algorithm}

\section{Datasets}

\subsection{Road Networks}
\label{sec:road_networks}

\begin{figure*}[t]
 		\centering
 		\includegraphics[width=1\textwidth]{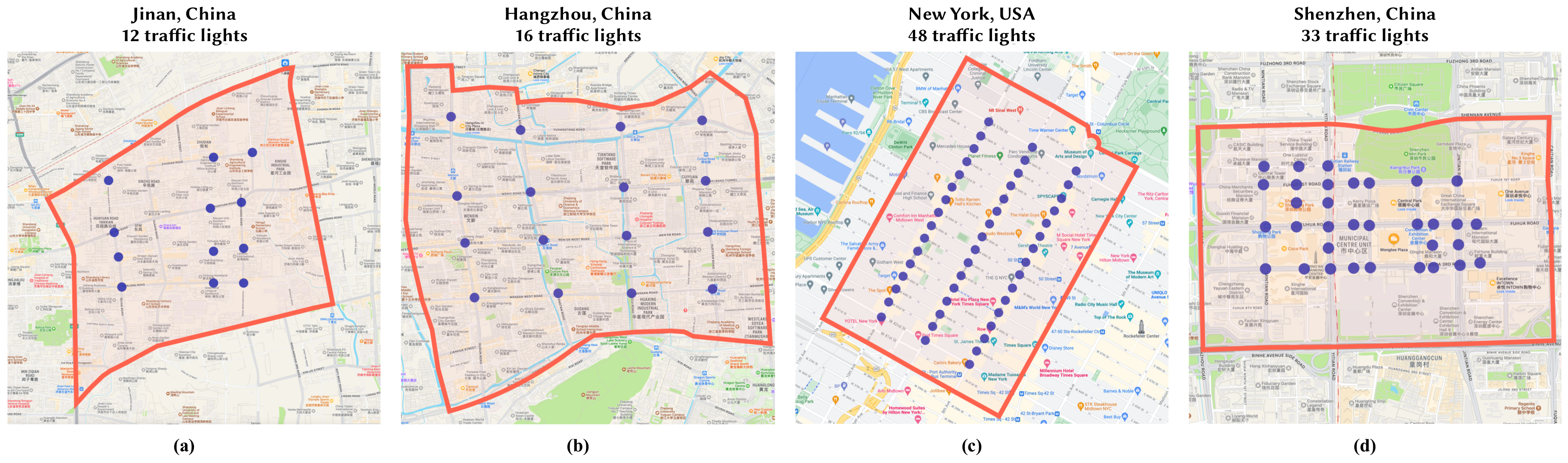} 
 		\vspace{-0.6cm}
 		\caption{The illustration of the road networks. The figures from left to right represent the road network of Jinan(China), Hangzhou(China), New York(USA) and Shenzhen(China), containing 12 ($4 \times 3$), 16 ($4 \times 4$), 48 ($16 \times 3$) and 33 (Non-grid) traffic signals respectively.}
 		\label{fig:road_network}
 		\vspace{-0.2cm}
\end{figure*}

The evaluation scenarios come from four real road network maps of different scales, including \textbf{Hangzhou} (China), \textbf{Jinan} (China), \textbf{New York} (USA) and \textbf{Shenzhen} (China), illustrated in Fig. \ref{fig:road_network}. The road networks and data of Hangzhou, Jinan and New York are from the public datasets\footnote{https://traffic-signal-control.github.io/}. The road network map of Shenzhen is made by ourselves which is derived from OpenStreetMap\footnote{The road network map and data of Shenzhen will be released to facilitate the future research.}. The road networks of Jinan and Hangzhou contain 12 and 16 intersections in $4 \times 3$ and $4 \times 4$ grids, respectively. The road network of New York includes 48 intersections in $16 \times 3$ grid. The road network of Shenzhen contains 33 intersections, which is not grid compared to other three maps.

\begin{figure*}[t]
    \centering
    \setlength{\abovecaptionskip}{5pt}
    \includegraphics[width=0.5\textwidth]{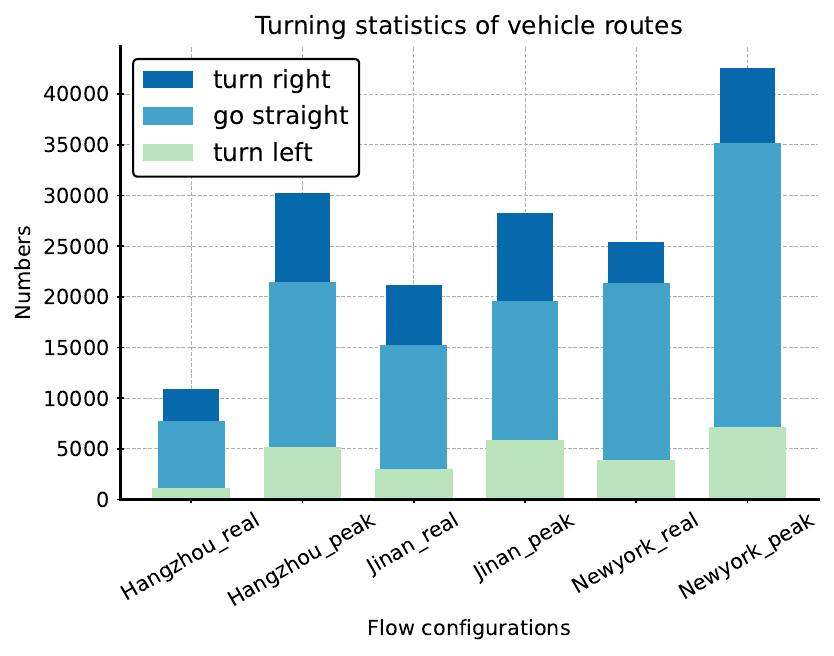}
    \caption{Turning statistics of vehicle routes.} 
    \vspace{-0.15cm}
    \label{fig:turning}
\end{figure*} 

\renewcommand\tabcolsep{5.0pt}  
\begin{table}[h] 
    \setlength{\belowcaptionskip}{6pt}
    \centering
    \caption{Arrival rate of real-world traffic dataset}
    \label{tab:data_statistics_1}
    \begin{tabular}{l l rrrr}
    \toprule
    \multirow{3}{*}{\textbf{Dataset}}
    & \multirow{3}{*}{\textbf{\# Intersections}}
    & \multicolumn{4}{c}{\textbf{Arrival rate (vehicles/300s)}} \\
    \cmidrule(lr){3-6}  
    & & \textbf{Mean} & \textbf{Std} & \textbf{Max} & \textbf{Min} \\
    \midrule
    \hangzhou & 16 (4 $\times$ 4) & 248.58 & 42.25 & 333 & 212 \\
    \jinan & 12 (4$\times$3) & 524.58 & 102.91 & 672 & 256 \\
    \newyork & 48 (16$\times$3) & 235.33 & 5.84 & 244 & 224 \\
    \shenzhen & 33 (Non-grid) & 147.92 & 79.35 & 255 & 22 \\
    \bottomrule
    \end{tabular}
    \vspace{-0.4cm}
\end{table}

\renewcommand\tabcolsep{2.5pt}\begin{table}[h]
    \centering
    \setlength{\belowcaptionskip}{6pt}
    \caption{Data statistics of synthetic traffic dataset}
    \label{tab:data_statistics_2}
\begin{tabular}{ccccc}
\toprule
    \textbf{Dataset}  & \textbf{Time}      & \begin{tabular}[c]{@{}l@{}} \textbf{Arrival rate}\\  \textbf{(vehicles/s)}\end{tabular} & \begin{tabular}[c]{@{}l@{}}\textbf{Incoming} \\ \textbf{vehicles} \end{tabular}  & \begin{tabular}[c]{@{}l@{}} \textbf{Accumulated} \\ \textbf{vehicles} \end{tabular}  \\ 
\midrule
    \multirow{6}{*}{\begin{tabular}[c]{@{}c@{}}\hangzhou /\\  \jinan /\\  \newyork /\\  \shenzhen \end{tabular}} 
    & 0-600     & 1.00   & 600    & 600     \\ 
    & 600-1200  & 0.25   & 150    & 750     \\ 
    & 1200-1800 & 4.00   & 2400   & 3150    \\ 
    & 1800-2400 & 2.00   & 1200   & 4350    \\ 
    & 2400-3000 & 0.2    & 120    & 4470    \\ 
    & 3000-3600 & 0.5    & 150    & 4770    \\ 
\bottomrule
\end{tabular}
\vspace{-0.3cm}
\end{table}

\renewcommand\tabcolsep{6pt}  

\begin{table}[h]
    \centering
    \small
    \setlength{\belowcaptionskip}{6pt}
    \vspace{-0.1cm}
    \caption{Statistics of turning frequency at intersections in all routes.}
    \label{tab:turning}
    \begin{tabular}{l  rr   rr  rr }
    \toprule
    \multirow{3}{*}{\textbf{\quad Model}}
    & \multicolumn{2}{c}{\textbf{Hangzhou}}
    & \multicolumn{2}{c}{\textbf{Jinan}} 
    & \multicolumn{2}{c}{\textbf{Newyork}} \\
    \cmidrule(ll){2-3} \cmidrule(ll){4-5} \cmidrule(ll){6-7} 
    & \textbf{real\quad } & \textbf{syn\_peak} & \textbf{real\quad } & \textbf{syn\_peak} & \textbf{real\quad } & \textbf{syn\_peak} \\
    \midrule
    \gr
    turn left & 1093 \scriptsize{(14\%)} & 5175 \scriptsize{(24\%)} & 3044 \scriptsize{(20\%)} & 5833 \scriptsize{(30\%)} & 3886 \scriptsize{(18\%)} & 7169 \scriptsize{(20\%)} \\
    \gr
    go straight & 6620 \scriptsize{(86\%)} & 16293 \scriptsize{(76\%)} & 12175 \scriptsize{(80\%)} & 13704 \scriptsize{(70\%)} & 17498 \scriptsize{(82\%)} & 27976 \scriptsize{(80\%)} \\
    turn right & 3184  & 8752 & 5972 & 8747  & 4021  & 7421 \\
    \bottomrule
    \end{tabular}
    \vspace{-0.2cm}
\end{table}

\subsection{Flow configurations}
\label{sec:flow_configurations}
We run the experiments under two traffic flow configurations: real traffic flow and synthetic traffic flow. The real traffic flow is real-world hourly statistical data with slight variance in vehicle arrival rates, as shown in Tab. \ref{tab:data_statistics_1}.  Since the real-world strategies tend to break down during bottleneck period (peak hour), to better evaluate the performances of traffic light control methods in the flat-peak-flat scenario, we use synthetic datasets, which have a more dramatic variance in vehicle arrival rates, as shown in Tab. \ref{tab:data_statistics_2}. A detailed description of traffic flow configurations is:

\begin{itemize}[leftmargin=15pt]
\item
\textbf{\realflow.} The traffic flows of \textbf{Hangzhou} (China), \textbf{Jinan} (China) and \textbf{New York} (USA) are from the public datasets, which are processed from multiple sources. The traffic flow of \textbf{Shenzhen} (China) is made by ourselves generated based on the traffic trajectories collected from 80 red-light cameras and 16 monitoring cameras in a hour. The data statistics are listed in Tab. \ref{tab:data_statistics_1}. 

\item
\textbf{\synflow.} The \synflow is a mixed traffic flow with a total flow of 4770 in one hour, to simulate a heavy peak. The arrival rate changes every 10 minutes, which is used to simulate the uneven traffic flow distribution in the real world, the details of the vehicle arrival rate and cumulative traffic flow are shown in Tab. \ref{tab:data_statistics_2}. 

\end{itemize}

\section{Evaluation Criteria}
Following existing studies \cite{wei2019colight,wei2019presslight,xiong2019learning,chen2020toward,zang2020metalight}, we use the \textbf{average travel time} to evaluate the performance of different methods for traffic signal control. The average travel time indicates the overall traffic situation in an area over a period of time. For a detailed definition of average travel time, see Section \ref{sec:preliminaries}. Since the number of vehicles and the origin-destination (OD) positions are fixed, better traffic signal control strategies result in less average travel time.

\section{Baselines}
\label{sec:baselines}
Our method is compared with the following two categories of methods: conventional transportation methods and RL methods\footnote{Some existing RL based traffic signal control methods, such as AttendLight \cite{oroojlooy2020attendlight} and SD-MaCAR \cite{guo2021urban}, evaluate their method under different experimental settings (e.g., road network or traffic flow), and the source codes are not available yet. Therefore, they are not compared in our experiments.}. Note that for a fair comparison all the RL methods are learned without any pre-trained parameters and the methods are evaluated under the same settings. The results are obtained by running the source codes\footnote{https://github.com/traffic-signal-control/RL\_signals}. All the baselines are run with three random seeds, and the mean is taken as the final result. The action interval is five seconds for each method, and the horizon is 3600 seconds for each episode. Specifically, the compared methods contain:

\subsection{Conventional methods}
\label{sec:conventional_methods}

\begin{itemize}[leftmargin=15pt]
\item
\textbf{\maxpressure \cite{varaiya2013max}} is a leading conventional method, which greedily chooses the phase with the maximum pressure. The pressure is defined as the difference of vehicle density between the incoming lane and the outgoing lane, and the vehicle density means the actual number of vehicles divided by the maximum permissible vehicle number.

\item
\textbf{\fixedtime \cite{koonce2008traffic}} with random offset \cite{roess2004traffic} executes each phase in a phase loop with a pre-defined span of phase duration, which is widely used for steady traffic.

\item
\textbf{\sotl \cite{cools2013self}} specifies a pre-defined threshold for the number of waiting vehicles on approaching lanes. Once the waiting vehicles exceeds the threshold, it will switch to the next phase.
\end{itemize}

\subsection{RL-based methods}
\label{sec:rl_based_methods}

\begin{itemize}[leftmargin=15pt]
\item
\textbf{\individualrl. \cite{wei2018intellilight}} Independent control is performed for each agent in multi-agent environment, each intersection is controlled by one agent. The replay buffer and network parameters are not shared, and the model update is independent. There is no information transfer between agents, and no neighbor information is considered. 

\item
\textbf{\metalight \cite{zang2020metalight}} is a value-based meta reinforcement learning method via parameter initialization, which is based on MAML \cite{finn2017model}. \metalight is originally a single-agent approach for meta-learning on multiple separate tasks. Here we extend it to a multi-agent scenario without considering neighbor information.

\item
\textbf{\presslight \cite{wei2019presslight}} combines the traditional traffic method \maxpressure \cite{varaiya2013max} with RL technology together. \presslight is a RL method that optimizes the pressure of each intersection.

\item
\textbf{\colight \cite{wei2019colight}} uses graph convolution and attention mechanism to model the neighbor information, and then further uses this neighbor information to optimize the queue length.

\item
\textbf{\generalight \cite{zhang2020generalight}} is a meta reinforcement learning method which uses generative adversarial network to generate diverse traffic flows and
uses them to build training environments.
\end{itemize}

\end{document}